\documentclass[lettersize,journal]{IEEEtran}
\usepackage{amsmath,amsfonts}
\usepackage{algorithmic}
\usepackage{algorithm}
\usepackage{array}
\usepackage[caption=false,font=normalsize,labelfont=sf,textfont=sf]{subfig}
\usepackage{textcomp}
\usepackage{stfloats}
\usepackage{url}
\usepackage{verbatim}
\usepackage{graphicx}
\ifCLASSOPTIONcompsoc
    \usepackage[caption=false, font=normalsize, labelfont=sf, textfont=sf]{subfig}
\else

\usepackage{cite}
\begin{document}

\title{A Pseudo-Boolean Polynomials Approach for Image Edge Detection }

\author{
    \IEEEauthorblockN{Tendai Mapungwana Chikake\IEEEauthorrefmark{1}, Boris Goldengorin\IEEEauthorrefmark{2}\IEEEauthorrefmark{3}\IEEEauthorrefmark{1}}

    \IEEEauthorblockA{
        \IEEEauthorrefmark{1,4}Department of Discrete Mathematics, Phystech School of Applied Mathematics and Informatics, Moscow Institute of Physics and Technology, Institutsky lane 9, Dolgoprudny, Moscow region, 141700, Russian Federation
    }

    \IEEEauthorblockA{
        \IEEEauthorrefmark{2} Department of Mathematics, New Uzbekistan University, Tashkent, 100007, Uzbekistan
    }

    \IEEEauthorblockA{
        \IEEEauthorrefmark{3} The Scientific and Educational Mathematical Center ``Sofia Kovalevskaya Northwestern Center for Mathematical Research'' in Pskov State University, Sovetskaya Ulitsa, 21, Pskov, Pskovskaya oblast', 180000, Russian Federation
    }
}

\maketitle

\begin{abstract}
    We introduce a novel approach for image edge detection based on pseudo-Boolean polynomials for image patches. 
    We show that patches covering edge regions in the image result in pseudo-Boolean polynomials with higher degrees compared to patches that cover blob regions. 
    The proposed approach is based on \textit{reduction} of polynomial degree and \textit{equivalence} properties of penalty-based pseudo-Boolean polynomials.
\end{abstract}


\section{Introduction}

\IEEEPARstart{I}{mage} edge detection refers to the process of identifying and locating boundaries between different regions or objects in an image \cite{umbaugh_digital_2011}.
It is usually performed through mathematical methods and in this article we propose a new approach based on pseudo-Boolean polynomials \cite{boros_pseudo_boolean_2002, goldengorinCellFormationIndustrial2013, albdaiwi_data_2011}.

The primary goal of edge detection is to identify the boundaries of objects within an image which can be achieved by detecting discontinuities in image intensity, colour, or texture. 
Over the years numerous edge detection techniques have been proposed ranging from classical gradient-based methods to more recent approaches based on machine learning and deep learning \cite{muntarinaNotesEdgeDetection2022a,jingRecentAdvancesImage2022}.

Image edge detection is mostly a processing step in myriads of computer vision and image processing tasks such as image segmentation, object recognition and image compression \cite{prasath_multiscale_2020}.
The process helps in simplifying image analysis through reduction of the amount of data that will be processed while at the same time preserving structural information about object boundaries \cite{Canny}.

In image segmentation tasks, edge detection is usually the first step used to identify the boundaries between different regions in an image.
It allows for demarcating the image into distinct regions which can then be used for further analysis or manipulation and eventually object segmentations \cite{article}.

Image edge detection is also used in object recognition tasks to improve the accuracy of object recognition algorithms by providing information about the shape and location of objects in an image \cite{786964}.
This information can be used to better identify objects and distinguish them from their backgrounds.

In image compression tasks, edge detection finds utility by reducing the amount of data needed to represent the image \cite{mishra_edge-aware_2021}. 
This is performed through identifying and removing the edges from an image so that the amount of data needed to represent the image is reduced, thereby resulting in smaller image file sizes.

Image edge detection is also used to enhance the visual quality of images by highlighting the boundaries between different regions in an image \cite{edge_enhancement}. 
This can make it easier to see and distinguish different objects or features in the image.

In industrial machine inspection low cost image detection can be used to monitor wear and tear on surfaces such as the use cases discussed in \cite{soleimani_applied_2021}.
There is intensive usage of image edge detection processes in biological research as shown in \cite{zhang_comprehensive_2022} where analysis of biological samples is performed on digital images instead of real samples.
This allows for obvious significant reductions in costs, logistical and storage problems.

Overall, image edge detection is a critical component of many computer vision applications such as autonomous vehicles, medical image analysis and security systems. In these applications, accurate edge detection is essential for accurately identifying and tracking objects in an image or across frames in a video \cite{jinImageEdgeEnhancement2022}.

Summing up all the use cases, we note that image edge detection is a fundamental building block for many other computer vision tasks and continues to be an active area of research \cite{prasath_multiscale_2020,mishra_edge-aware_2021,soleimani_applied_2021,zhang_comprehensive_2022,frsip.2022.826967,jiang2020emphcmsalgan,9887911}.

As far as we are aware in this article we present a uniquely new approach. Our approach is based on the degree property of pseudo-boolean polynomials formulated on patches extracted from a digital image matrix. Preliminary study cases that we present in this paper is based on properties of pseudo-Boolean polynomials as shown in \cite{chikake_dimensionality_2023, chikake_pseudo-boolean_2023}
We demonstrate that our approach is simple, fast and adaptable for many cases.

\section{Related methods}

\noindent There have been many approaches to image edge detection methods over the years, with the earliest methods dating back to the 1960s and 1970s \cite{frsip.2022.826967} and recent ones \cite{jingRecentAdvancesImage2022}. 

We can divide the techniques used in image edge detection into three groups with regard to their times of publication: Early Edge Detection Techniques, Recent Edge Detection Techniques and Modern Edge Detection Techniques.

\subsection{Early Edge Detection Techniques}

Early Edge Detection Techniques involve identifying regions where rapid changes in signal intensity values occur. This group of techniques is also known as gradient-based edge detection techniques and the popular methods include the Sobel\cite{frsip.2022.826967}, Prewitt \cite{prewitt_object_1970}, and Canny \cite{Canny} operators.

One of the earliest and most well-known methods is the Sobel operator, which uses a combination of gradient operators to detect edges in an image \cite{frsip.2022.826967}. 
It computes the gradient of an image using two ${3 \times 3}$ convolution kernels, one for the horizontal gradient ${G_x}$ and one for the vertical gradient ${G_y}$. 
The magnitude of the gradient ${G}$ is then calculated as the square root of the sum of the squares of ${G_x}$ and ${G_y}$, and the edge direction (${\theta}$) is computed as the arctangent of the ratio of ${G_y}$ to G \cite{sobel_isotropic_2015}.
The Sobel operator has been widely used in many applications due to its computational efficiency and relatively simple implementation, but it is sensitive to noise and may produce false edges in the presence of noise \cite{mittal_efficient_2019}.

The Prewitt operator \cite{prewitt_object_1970} is another gradient-based edge detection technique that is similar to the Sobel operator. 
The main difference between the two methods is the choice of convolution kernels used to compute the horizontal and vertical gradients. 
The Prewitt operator uses simpler kernels, which result in slightly less accurate edge detection compared to the Sobel operator. 
However, the Prewitt operator is less sensitive to noise, making it more suitable for noisy images.

The Canny method\cite{Canny}, is relatively old, but it is still competitive and extensively used as is or in improved versions in recent results \cite{liu_image_2017}.
The Canny operator consists of several steps including noise reduction using a Gaussian filter, gradient computation using the Sobel operator, non-maximum suppression to thin the edges, and hysteresis threshold processing to eliminate weak edges. 
It is widely regarded as one of the best edge detection techniques in terms of accuracy and robustness to noise. 
However, it is more computationally expensive than the Sobel and Prewitt operators and requires the selection of several parameters, such as the size of the Gaussian filter and the threshold values for hysteresis threshold processes.
It has been shown in \cite{liu_image_2017} that the Canny method is adaptable to various environments because its parameters allow it to be tailored to recognition of edges of differing characteristics depending on the particular requirements of a given implementation. 
The optimized Canny filter is recursive, and can be computed in a short, fixed amount of times.
However, the implementation of the Canny operator does not give a good approximation of rotational symmetry and therefore gives a bias towards horizontal and vertical edges as shown in \cite{liu_image_2017}. 
Our approach makes use of the Gaussian filter preprocessor, hysteresis threshold processing and finally calculating a pseudo-Boolean polynomial in normal and transposed patches on the image matrices.
We select those regions where the resulting pseudo-Boolean polynomials have higher degrees and classify those regions as either edges or blobs.
Consequently, our approach can be considered a member of gradient-based image edge detection techniques. 

From an indirect perspective we can look at aggregation of possible distinct objects in image data in terms of blob extraction. 
The regions outside these blobs can be interpreted as edge regions.

This group of methods is common in the group of Recent Edge Detection Techniques.

\subsection{Recent Edge Detection Techniques}

In this type of image edge detection techniques, popular methods are similar to Laplacian-based edge detection techniques which rely on the second-order derivative of the image intensity to identify edges. The Laplacian of an image is a scalar quantity that measures the curvature of the intensity surface and is computed as the sum of the second order partial derivatives in the ${x}$ and ${y}$ directions \cite{Laplacian_of_Gaussian}. 

Such popular solutions have an underlying combination of methods among the popular blob extraction techniques which include Laplacian of Gaussian \cite{Laplacian_of_Gaussian}, Difference of Gaussian \cite{assirati_performing_2014}, or Determinant of a Hessian \cite{li_pattern_2014}, among others.

Laplacian of Gaussian is a blob extraction method which determines the blobs by using the Laplacian of Gaussian filters \cite{Laplacian_of_Gaussian}. 
The Laplacian is a 2-D isotropic measure of the 2nd spatial derivative of an image which highlights regions of rapid intensity change and is therefore often used for edge detection \cite{Laplacian_of_Gaussian}. 
Researchers often apply the Laplacian after an image smoothing method with something approximating a Gaussian smoothing filter in order to reduce its sensitivity to noise. 

The Difference of Gaussian method determines blobs by using the difference of two differently sized Gaussian-smoothed images and follows generally most of the concept of the Laplacian of Gaussian \cite{Laplacian_of_Gaussian}. 

The aggregation of possible segmentation regions is achieved by finding blobs using the maximum in the matrix of the Hessian determinant \cite{li_pattern_2014}.

On the other hand, our approach aggregates equivalent regions by assigning regions that result in pseudo-Boolean polynomials with lower degree as blob regions and those with higher degrees as edges.
Regions described by rapidly distinct signal distributions in the pixel array result in high-degree pseudo-Boolean polynomials which indicate contour regions.
This property allows us to extract larger and fewer spatial regions of interest in an image making our approach superior to the current blob aggregation methods.

\subsection{Modern Edge Detection Techniques}

In recent years, there has been a growing interest in using deep learning and statistical techniques for image edge detection. 
In this group we also include methods that do not necessarily involve learning patterns in images using deep learning models but also recent and novel methods that use different approaches to the classical and Laplacian-based methods. 
One example is the method introduced in \cite{prasath_multiscale_2020} that uses multiscale gradient maps which enables better edge localizations, robust edge maps and local thresholding using Fisher information.
Fisher information (FI) is a non-parametric measure determined by a probability density function used for finding the probability of a grey level in a greyscale image \cite{prasath_multiscale_2020}.
The method in \cite{prasath_multiscale_2020} shows competitive results compared to classical methods like Canny.
In this article we also provide our visual comparison against this method.
Another modern edge detection technique based on generalized type-2 fuzzy logic is discussed in \cite{melin_edge-detection_2014}. This method belongs to the family of techniques that uses fuzzy logic algorithms and genetic algorithms shown in \cite{Setayesh}.

Deep learning based methods typically involve training a deep neural network to extract the features important for edge detection on the given dataset of images. 
These methods have been shown to be highly effective with state-of-the-art performance on many edge detection benchmarks \cite{muntarinaNotesEdgeDetection2022a}.
One example of a recent paper on this topic is ``Red-Green-Blue-Dense (RGB-D) Salient Object Detection with Cross-View Generative Adversarial Networks'' by \cite{jiang2020emphcmsalgan}. 
\cite{jiang2020emphcmsalgan} proposes a deep learning-based method for edge detection that uses a novel cross-modality Saliency Generative Adversarial Network to learn the important features of saliency detection.
Saliency essentially refers to the important edges of objects found in the image \cite{tjoaQuantifyingExplainabilitySaliency2022}. 

Another example is published in \cite{9887911}. 
The authors propose a method for multiscale edge detection that uses a deep feature embedding network to learn a set of features from the input image. 
These features are then used to detect edges at different scales for industrial inspection of textured surfaces—in the form of visual inspection \cite{9887911}.

Despite these recent high performing deep learning based methods the traditional methods are still in active use and are desired for their simplicity and computational efficiencies. 
Our approach is based on transforming data into a pseudo-Boolean polynomial and cannot be classified within any machine learning methods. 
Our approach can thus be considered as a class of deterministic (algebraic) methods that is completely new, i.e. unrelated to any of the well known statistical approaches.

\section{Pseudo-Boolean Polynomials for image edge detection}

\noindent In mathematics a pseudo-Boolean polynomial (pBp) is a
mapping ${f: \mathbf{B^n} \to \mathbb{R}}$. Here ${\mathbf{B} = \{0, 1\}}$ is a Boolean domain and ${n}$ is a natural number called the degree of the pBp.  \cite{boros_pseudo_boolean_2002}

In \cite{goldengorinCellFormationIndustrial2013} a fundamental reduction and equivalence relation induced by a pBp are presented.
This relation can be described in terms of a polyhedron representing a wide set of equivalent instances. 

We specify and adjust pBp approach to image processing leading to an alternative aggregated representation of digital image data. 
This aggregated data representation allows us to analyse image data from a different perspective.

Instead of looking at the image pixel data as mere colour intensities we are now able to see the pixel values as numerical measures of colour intensities in respective spatial regions. 

By computing pseudo-Boolean polynomials on image patches we are able to achieve a compact and illustrative polynomial on each region across the image matrix.
In this context an image patch refers to a small rectangular piece of an image that can be seen as a subarray of the original image array where each element is a pixel value positioned in some spatial region.

Having image regions represented in terms of pseudo-Boolean polynomials allows us to analyse regions on the image patch by comparing pBp properties like \textit{degree}, \textit{equivalence} and \textit{polyhedra} of equivalent instances \cite{albdaiwi_data_2011}.

In terms of the edge detection task two properties, namely polynomial degree and truncation \cite{goldengorinCellFormationIndustrial2013} provide analytical tools to solve our problem.

We are given an image patch represented by an ${m \times n}$ matrix which we treat as an information cost matrix ${C}$. 
Our goal is to determine if pixels in a column lie in the same spatial region by comparing their pixel intensities.
The rows ${I = {1, 2, \dots, m}}$ represent spatial regions and columns ${J = {1, 2, \dots, n} }$ are pixel positions while matrix ${C = [c_{ij}]}$ introduces pixel intensities of each ${j \in J}$ pixel lying in the spatial region ${i \in I}$.
In this context a spatial region in the image is a set of pixels that share common characteristics or properties such as colour, intensity or texture \cite{1211447}.

Following the formulation of pseudo-Boolean polynomials as explained in \cite{goldengorinCellFormationIndustrial2013, albdaiwi_data_2011} we argue that for a pixel column ${j}$, we consider ${\Pi^j = (\pi_{ij}, \dots, \pi_{mj} )}$ a permutation of ${1, \dots, m}$ such that ${c_{\pi_{ij}j} \le c_{\pi_{kj}j}}$ if ${}i < k$ for all ${i, k \in {1, \dots, m}}$.
Further ${\Delta^j = (\delta_{1j}, \dots, \delta_{mj})}$ are differences with ${\delta_{1j} = c_{\pi_{1j}j}, \text{and}\, \delta_{rj} = c_{\pi_{rj}j} - c_{\pi_{(r-1)j}j}}$ for ${r = 2 \dots, m}$ and argue that a pixel's intensity at ${\pi_{1j}}$ lies in a spatial region ${i}$ with a representational cost of ${c_{\pi_{1j}j}}$ in the column.
If the next pixel in the column lies in the same spatial region ${i}$ then the representational cost ${c_{\pi_{2j}j}}$ should be zero unless the neighbouring pixels lie in different spatial regions.

In \cite{goldengorinCellFormationIndustrial2013, albdaiwi_data_2011} the authors define an \textit{m}-vector ${\mathbf{y} = (y_1, \dots, y_m)}$ with ${y_i = 0}$, if a pixel lies in the spatial region ${i}$ and 1, otherwise.
The representational total cost for spatial positioning of a pixel at ${j}$ is given by the pseudo-Boolean polynomial 
\begin{equation}
    \label{eq:1}
    f^{i}_C(\mathbf{y}) = \delta_{1j} + {\sum^m_{k=2} \delta_{k, j}} \cdot {\prod_{r=1}^{k-1} y_{\pi_{rj}}}
\end{equation}

If a patch overlaps regions of contrasting information (i.e. overlapping an edge) in an image then the computed pseudo-Boolean polynomial has a larger degree in comparison to one that lies over a blob region.

The above mentioned differences reflecting pixel intensities in blob regions are usually equivalent or have very small differences and similar costs cancel each other out.
In an ideal case a blob has an unchanging colour with intensities canceling each other out in the column resulting in a constant term within pBp. 
However, natural images rarely have these ideal blobs: a situation that requires us to set a truncating degree ${p}$ of pBp which serves as a threshold classifier for distinctly different spatial regions.

Truncation of the pseudo-Boolean polynomial serves as the edge/blob classification.
We represent the classification by truncation as:

\begin{equation}
    \label{eq:2}
    f(d, p) =
    \begin{cases}
        \text{edge} & \text{if}\, d > p, \\
        \text{blob} & \text{otherwise}
    \end{cases}
  \end{equation}
  where ${p}$ is the truncating value which determines whether an image patch with pseudo-Boolean polynomial of degree ${d}$ lying over an edge or over a blob region.

To implement the edge detection process we introduce two parameters: \textit{image patch size} and the \textit{truncated polynomial degree} ${p}$ depending on the image complexity and edge detection sensitivity.

The smaller the image patch size, the finer the edge detection and vice-versa while the \textit{truncated degree} ${p}$ controls the significance of the edges. 
A higher ${p}$ value discourages less significant edges while a lower value encourages flagging of the slightest contours.

The \textit{image patch size} represents the rows and columns of the image patch which determine the height and width in pixels, respectively.
For example, a ${3 \times 3}$ image patch has 3 rows and 3 columns and covers a ${3 \times 3}$ pixel area in the image.
To reduce the computational complexity (CPU times, memory and space complexity of algorithms) we operate on greyscale image patches that have their pixel ranges reduced by multiple thresholding \cite{aroraMultilevelThresholdingImage2008}.

We also output two pseudo-Boolean polynomials, ${E\, \text{and} \, E'}$ from normal and transposed image patches, respectively.  
Depending on the required sensitivity of the edge detections we set a constraint satisfying both pseudo-Boolean polynomials (\ref{eq:2}).

Using an example to illustrate the formulation process we take a ${4 \times 5}$ patch from an image in the form of matrix 

\[
C = 
    \begin{bmatrix}
        8 & 8 & 8 & 5\\
        12 & 7 & 5 & 7\\
        18 & 2 & 3 & 1\\
        5 & 18 & 9 & 8\\
    \end{bmatrix}
\]

After ordering the entries in each column we create a permutation matrix (${\Pi}$) which is an index ordering of column entries in a non-decreasing order \cite{goldengorinCellFormationIndustrial2013, albdaiwi_data_2011}.

\[
	\Pi = 
    \begin{bmatrix}
        4 & 3 & 3 & 3\\
        1 & 2 & 2 & 1\\
        2 & 1 & 1 & 2\\
        3 & 4 & 4 & 4\\
    \end{bmatrix}
\]

Using ${\Pi}$, we sort the initial cost matrix ${C}$ 

\[
	\text{sorted}\, C = 
    \begin{bmatrix}
        5 & 2 & 3 & 1\\
        8 & 7 & 5 & 5\\
        12 & 8 & 8 & 7\\
        18 & 18 & 9 & 8\\
    \end{bmatrix}
\]

and derive the ${\Delta C}$ matrix showing the differences between two neighbouring entries and keeping the smallest one in the first row of ${\Delta C}$ 

\[
	\Delta C  = 
    \begin{bmatrix}
        5 & 2 & 3 & 1\\
        3 & 5 & 2 & 4\\
        4 & 1 & 3 & 2\\
        6 & 10 & 1 & 1\\
    \end{bmatrix}
\]

Using the ${\Pi}$ matrix we calculate the terms’ matrix y to represent the sorted matrix ${C}$ depending on ${0, 1}$ values of Boolean variables ${y_j}$

\[
	\mathbf{y} =  
	\begin{bmatrix}
        \\
        y_{4} & y_{3} & y_{3} & y_{3}\\
        y_{1}y_{4} & y_{2}y_{3} & y_{2}y_{3} & y_{1}y_{3}\\
        y_{1}y_{2}y_{4} & y_{1}y_{2}y_{3} & y_{1}y_{2}y_{3} & y_{1}y_{2}y_{3}\\
    \end{bmatrix}
\]

and derive the resulting monomials of pBp

\[
	E = \begin{bmatrix}
        5 & 2 & 3 & 1\\
        3y_{4} & 5y_{3} & 2y_{3} & 4y_{3}\\
        4y_{1}y_{4} & 1y_{2}y_{3} & 3y_{2}y_{3} & 2y_{1}y_{3}\\
        6y_{1}y_{2}y_{4} & 10y_{1}y_{2}y_{3} & 1y_{1}y_{2}y_{3} & 1y_{1}y_{2}y_{3}\\       
	\end{bmatrix}
\]

from which we perform local aggregation by reduction of similar monomials and get a compact representation of the initial instance as

\[
    E = \begin{bmatrix}
        0 & 0 & 11\\
        0 & 11y_{3} & 3y_{4}\\
        2y_{1}y_{3} & 4y_{2}y_{3} & 4y_{1}y_{4}\\
        0 & 12y_{1}y_{2}y_{3} & 6y_{1}y_{2}y_{4}\\
    \end{bmatrix}
\]

By counting the terms in the resulting pseudo-Boolean polynomial 

\begin{multline}
  B(y) = 11 + 11y_{3} + 3y_{4} + 2y_{1}y_{3} + \\
  4y_{2}y_{3} + 4y_{1}y_{4} + 12y_{1}y_{2}y_{3} + 6y_{1}y_{2}y_{4}
\end{multline}
we observe that for this particular example we managed to reduce the space complexity by ${50\%}$ since ${C}$ contains ${16}$ entries while ${B(y)}$ has just ${8}$ monomials (3).

When pseudo-Boolean polynomials are reduced to their smallest instance an equivalence property can be applied from initially different instances. 
This equivalence enhances our pattern recognition analysis and assists us in grouping structurally similar regions across the image matrix.
Below are examples of cost matrices, which are initially different but converge into similar reduced instances.

\[
	\begin{bmatrix}
		138 & 138 & 138 & 136 \\
		139 & 139 & 138 & 137 \\
		142 & 141 & 139 & 138 \\
		142 & 140 & 139 & 138 \\
	\end{bmatrix}
\]

and 

\[
	\begin{bmatrix}
		136 & 136 & 138 & 140 \\
		138 & 137 & 138 & 140 \\
		140 & 139 & 140 & 141 \\
		139 & 139 & 140 & 141 \\
	\end{bmatrix}
\]

can be reduced to 

\[
	\begin{bmatrix}
		550\\
		3y_1\\
		6y_1y_2\\
		1y_1y_2y_4\\
	\end{bmatrix}
\]

Using the argument that regions of continuous colour intensities result in terms that cancel each other out and therefore fewer terms and/or terms with lower degrees ${d}$, we can create an edge/blob classifier as shown in (\ref{eq:2}).

Adjusting ${p}$ allows us to control the sensitivity of edge detection.

\section{Experimental Setup}

\noindent To demonstrate the process of image edge detection using pseudo-Boolean polynomial formulation on image patches we consider both simple and complex images. 

\begin{figure}[h]
  \centering
    \subfloat[\label{1a}]{%
        \includegraphics[width=0.45\linewidth]{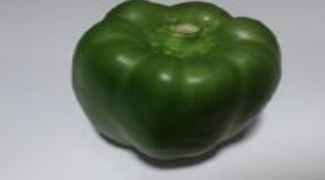}}
      \hfill
    \subfloat[\label{1b}]{%
          \includegraphics[width=0.45\linewidth]{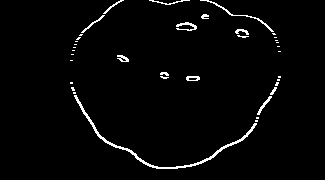}}
      \\
  \caption{(a) Input image of a pepper, (b) edge detection output is exposed using pseudo-Boolean polynomial formulation on the pepper image.}
  \label{fig:pepper_pbp}
\end{figure}

\begin{figure}[h]
    \centering
      \subfloat[\label{1c}]{%
          \includegraphics[width=0.45\linewidth]{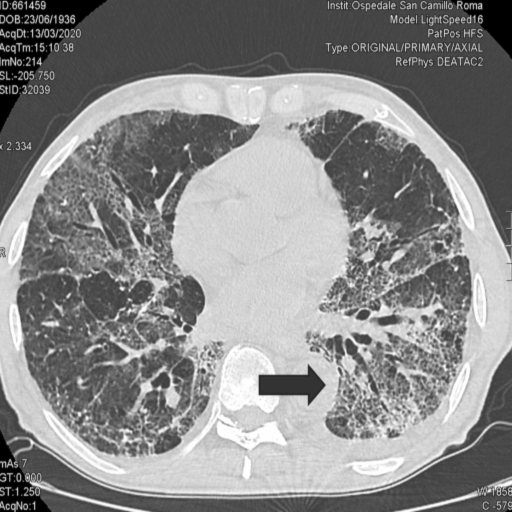}}
        \hfill
      \subfloat[\label{1d}]{%
            \includegraphics[width=0.45\linewidth]{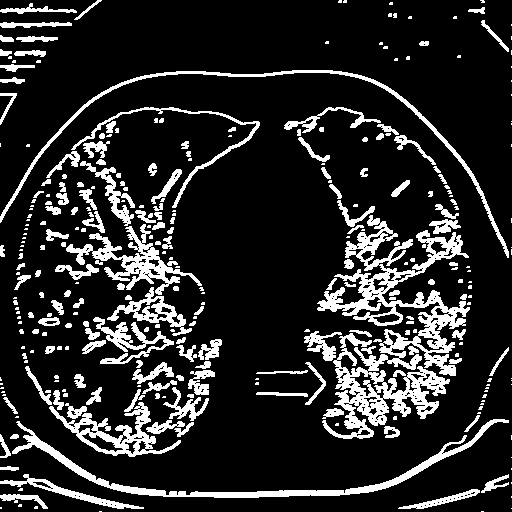}}
        \\
    \caption{(a) Input image of a human body cross-section scan, (b) edge detection output mask using pseudo-Boolean polynomial formulation }
    \label{fig:cross_section_scans_pbp}
  \end{figure}

Fig.~\ref{fig:pepper_pbp} and Fig.~\ref{fig:cross_section_scans_pbp} present simple images with limited colour shades and the output edge is exposed after applying the edge detection using pseudo-Boolean polynomial formulation.

Given an image with fewer colours and contours as shown in Fig.~\ref{fig:pepper_pbp} and Fig.~\ref{fig:cross_section_scans_pbp} we extract patches of significantly smaller sizes, e.g. ${6 \times 6}$, and set ${p = 3}$.
From these patches we apply pseudo-Boolean polynomial formulation in both vertical and horizontal orientations and take the maximum polynomial degree ${d = \max\{deg(f_x, f_y)\}}$ value of either one.

We group the patch regions into a binary set ${S = \{\text{Blob}, \text{Edge}\}}$ based on the pseudo-Boolean polynomial degree.

Patches with pseudo-Boolean polynomial degree ${d < p}$ are considered \textbf{Blobs} and \textbf{Edge}, otherwise.

For illustration purposes lets consider ${4}$ image patches labeled ${C_a, C_b, C_c, C_d}$.  
These example image patches are arbitrally selected to showcase the different characteristics of resulting pseudo-Boolean polynomials and how we use them to determine if a given patch is extracted from an edge or a blob region.

Patches extracted from blob regions on the image matrix can have constant pixel values ${x \in [0, 255]}$, like 
\[
    C_a = \begin{bmatrix}
        99 & 99 & 99 & 99\\
        99 & 99 & 99 & 99\\
        99 & 99 & 99 & 99\\
        99 & 99 & 99 & 99\\
    \end{bmatrix}
\]
are guaranteed to converge to a constant (zero-degree pseudo-Boolean polynomial) 

\[
    E_a = \begin{bmatrix}
        396\\
    \end{bmatrix}
\]
while some with varied pixel values cancel each other out. 
An example of an image patch with varied pixel values  
\[
    C_b = \begin{bmatrix}
        254 & 254 & 19 & 84\\
        254 & 254 & 19 & 84\\
        254 & 254 & 19 & 84\\
        254 & 254 & 19 & 84\\
    \end{bmatrix}
\]
and its respective pBp is a constant
\[
    E_b = \begin{bmatrix}
        611\\
    \end{bmatrix}
\] 
indicating that no edges are forming in the vertical sense. 

However, when we transpose this example patch ${C_b}$ to
\[
    C_b' = \begin{bmatrix}
        254 & 254 & 254 & 254 \\
        254 & 254 & 254 & 254 \\
        19  &  19 & 19  & 19  \\
        84  &  84 & 84  & 84  \\
    \end{bmatrix}
\]

a different pseudo-Boolean polynomial with a quadratic polynomial degree results in ${76 + 260y_{3} + 680y_{3}y_{4} }$ which indicates that there is an edge in the patch, but only in the horizontal direction.

Thus we have shown that our approach is flexible in detecting both vertically and horizontally orientated edges. 

A high-degree pseudo-Boolean polynomial results when we process an image patch that lie over a contour region. For example, 
\[
    C_c= \begin{bmatrix}
        254 & 254 & 6 & 17\\
        254 & 254 & 6 & 17\\
        254 & 254 & 6 & 17\\
        254 & 254 & 6 & 123\\
    \end{bmatrix}
\]
and 
\[
   C_d =  \begin{bmatrix}
        254 & 254 & 6 & 17\\
        254 & 254 & 6 & 123\\
        254 & 254 & 6 & 123\\
        254 & 254 & 6 & 123\\
    \end{bmatrix}
\]
both converge to 
\[
    E_{c,d} = \begin{bmatrix}
        531\\
        106y_{1}y_{2}y_{3}\\
    \end{bmatrix}
\]

In ${C_c\, \text{and}\, C_d}$ we can observe varied entries in the far-right column which indicates entries that lie in distinct spatial regions while the other columns have constant pixel values indicating similar spatial regions.
The resulting pBp for ${C_c\, \text{and}\, C_d}$ is a 3rd-degree pseudo-Boolean polynomial because of the varied pixel entries in the far-right column.
Thus patches ${C_c\, \text{and}\, C_d}$ are classified as lying over edge regions if we choose the truncated degree ${p = 2}$ because their pBp degree ${d = 3 }$ and ${d > p}$ holds true.

\begin{figure}[h]
    \centering
    \includegraphics[width=1\linewidth]{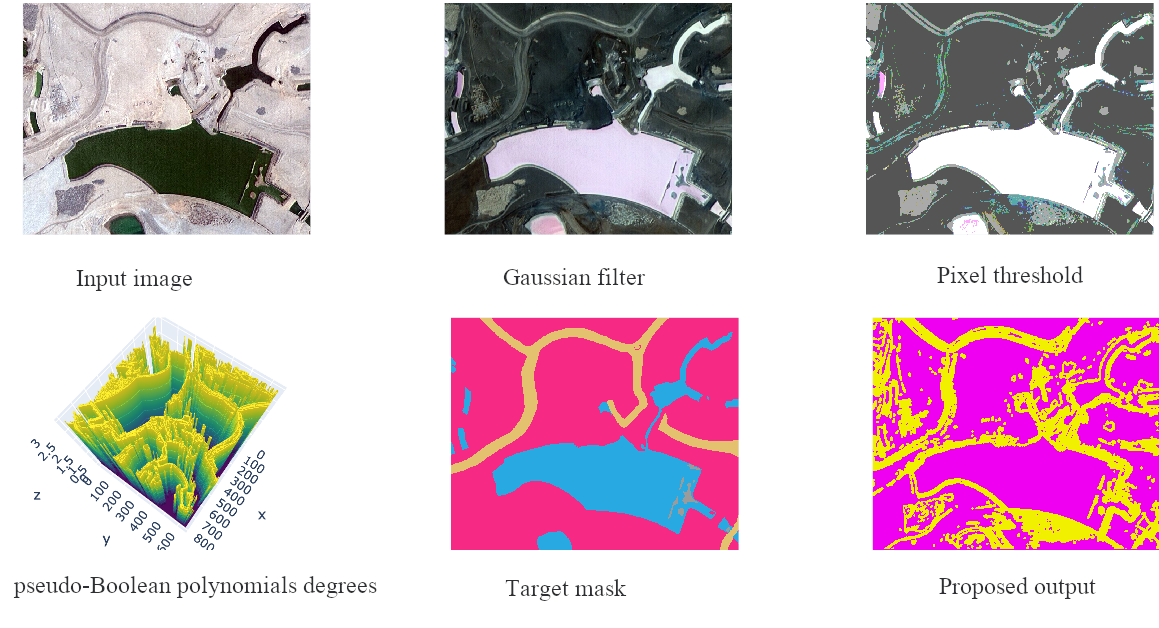}
    \caption{Edge detection process using pseudo-Boolean polynomials  }
    \label{fig:landscape__pbp}
\end{figure}

Natural images tend to have smooth transitions of pixel values for neighbouring pixels due to anti-aliasing in representation of image data, thereby reducing the chances of neighbouring pixel patches converging into equivalent groups and consequently coarse edges.
For this reason we use the Gaussian filter as a preprocessing step followed by a pixel set aggregation step (i.e. image colour reduction) which is essentially a multivalued threshold processing \cite{aroraMultilevelThresholdingImage2008}. 
These steps are shown in Fig.~\ref{fig:landscape__pbp} including a surface plot of the output pseudo-Boolean polynomials' degrees. 
In Fig.~\ref{fig:landscape__pbp} a target mask is also shown to provide a visual comparison of edge and blob annotations between a manual processed output (Target mask) and the pBp processed output (Proposed output).

\begin{figure}[h] 
    \centering
      \subfloat[\label{2a}]{%
          \includegraphics[width=0.3\linewidth]{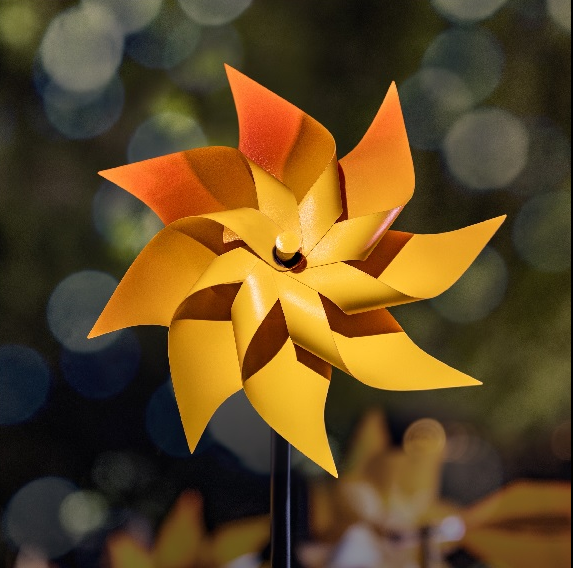}}
      \hfill
      \subfloat[\label{2b}]{%
        \includegraphics[width=0.3\linewidth]{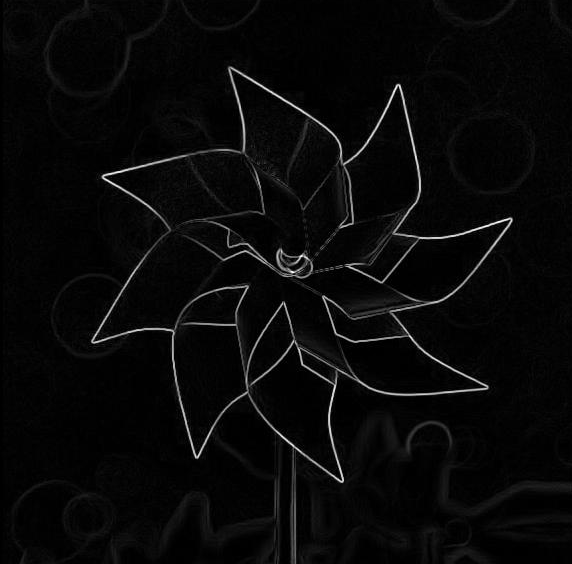}}
      \hfill
      \subfloat[\label{2c}]{%
            \includegraphics[width=0.3\linewidth]{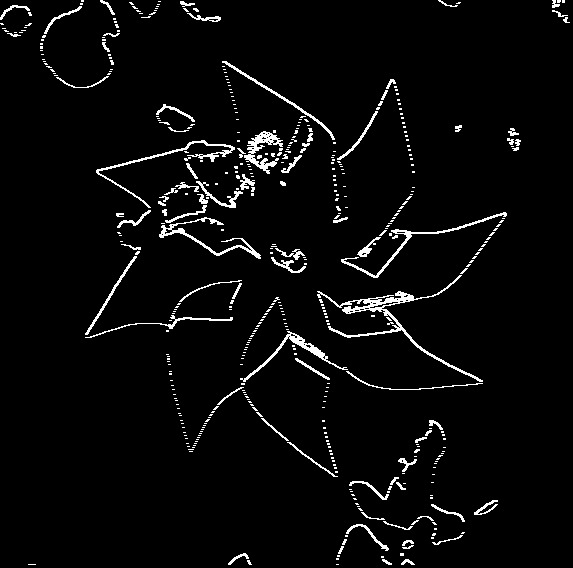}}
      \\
      \caption{(a) Input image, (b) edge detection output mask using a Sobel operator, (c) edge detection output mask using pseudo-Boolean polynomial formulation}
    \label{fig:flower_pbp}
\end{figure}

Fig.~\ref{fig:flower_pbp} shows a visual comparison between the Sobel operator and our approach. 
Our approach misses some edges compared to the edges detected by the Sobel operator. 
This limitation is attributed to course threshold parameters which control the sensitivity of our approach.

The above examples show that the choice of preprocessing parameters: Gaussian filter size, threshold processing ranges and edge/blob classifier degree ${p}$ affect the performance quality of our approach. 
To improve the performance quality we start the threshold preprocess by grouping ranges of pixels into \textit{pixel sets} by pixelation or colour palette reduction where sizes depend on the variance of pixel distribution in a given image.

These pixel-groups allow our approach to process smaller pixel intensity ranges like ${[0, 10]}$ instead of ${[0, 255]}$. 
Consequently this improves the computing runtime and computing memory usage.

\section{Results and Discussion}

\noindent As shown in the above examples our approach compared to Canny and Sobel methods outperforms them in quality since it is tunable to directly express the fineness/coarseness of edge detection. 
This is made possible by adjusting the edge/blob threshold ${p}$ and the patch size. 

\begin{figure}[h]
  \centering
\subfloat[\label{3a}]{%
     \includegraphics[width=0.5\linewidth]{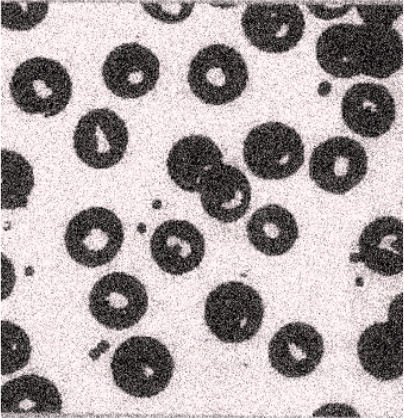}}
  \hfill
\subfloat[\label{3b}]{%
      \includegraphics[width=0.5\linewidth]{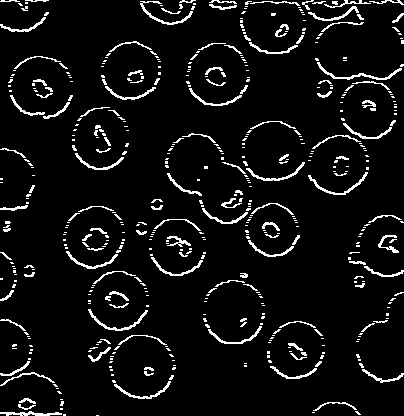}}
  \\
\subfloat[\label{3c}]{%
      \includegraphics[width=0.5\linewidth]{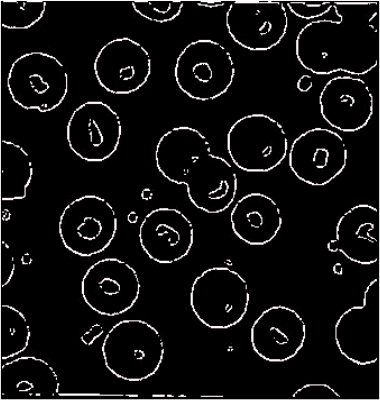}}
  \hfill
\subfloat[\label{3d}]{%
      \includegraphics[width=0.5\linewidth]{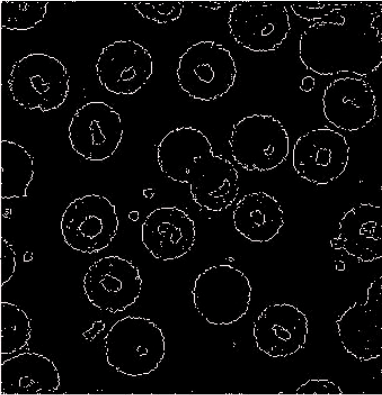}}
\caption{(a) Input image of cells in \cite{zhang_jin-yu_edge_2009}, (b) edge detection output mask using pseudo-Boolean polynomial formulation (c) Canny based edge detection output mask as reported in \cite{zhang_jin-yu_edge_2009}, (d) Sobel based edge detection output mask as reported in \cite{zhang_jin-yu_edge_2009}.}
\label{fig:pbp_canny_sobel}
\end{figure}

We show a visual comparison of edge detection using our pseudo-Boolean polynomials method against the Canny and Sobel methods in Fig.~\ref{fig:pbp_canny_sobel}.
In the comparison, we observe that edges detected by our approach are more continuous in comparison than the ones detected by the Sobel and Canny operators.

\begin{figure}[h]
  \centering
    \subfloat[\label{4a}]{%
        \includegraphics[width=0.3\linewidth]{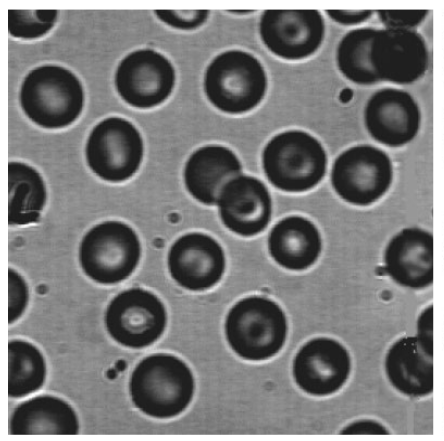}}
    \subfloat[\label{4b}]{%
        \includegraphics[width=0.29\linewidth]{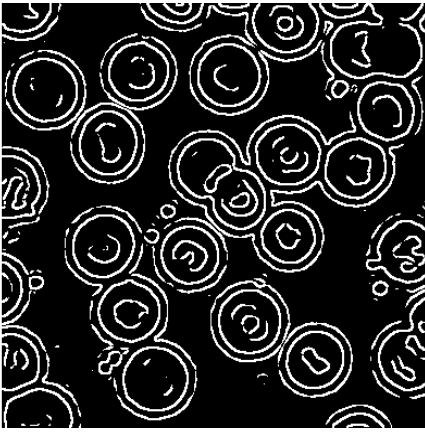}}
    \subfloat[\label{4c}]{%
        \includegraphics[width=0.295\linewidth]{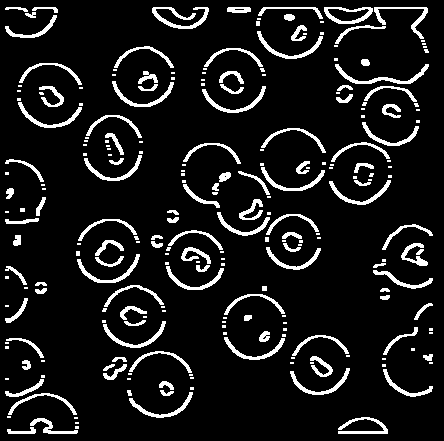}}
  \\
    \subfloat[\label{4d}]{%
        \includegraphics[width=0.3\linewidth]{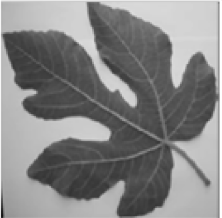}}
    \subfloat[\label{4e}]{%
        \includegraphics[width=0.3\linewidth]{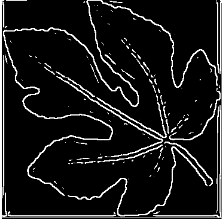}}
    \subfloat[\label{4f}]{%
        \includegraphics[width=0.3\linewidth]{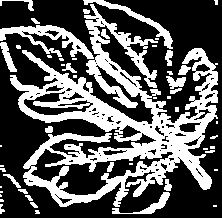}}
    \\
    \subfloat[\label{4g}]{%
        \includegraphics[width=0.3\linewidth]{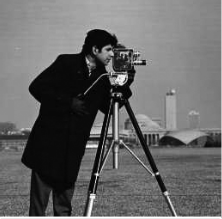}}
    \subfloat[\label{4h}]{%
        \includegraphics[width=0.3\linewidth]{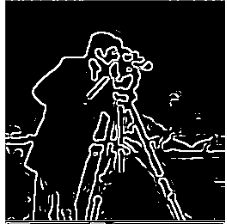}}
    \subfloat[\label{4i}]{%
        \includegraphics[width=0.3\linewidth]{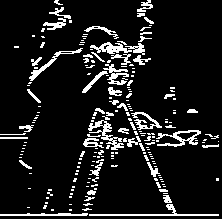}}

\caption{(a), (d), (g)  Input images in \cite{prasath_multiscale_2020}, (b), (e), (f) edge detection output mask using \cite{prasath_multiscale_2020}'s method (c), (f), (i) edge detection output mask using pseudo-Boolean polynomials-based method}
\label{fig:pbp_prasath}
\end{figure}

Our approach, see  e.g. Fig.~\ref{fig:pbp_prasath}, captured more edges in a
visual comparison with the recently published method in \cite{prasath_multiscale_2020}.

\begin{figure}[h]
    \centering
      \subfloat[\label{5a}]{%
          \includegraphics[width=0.3\linewidth]{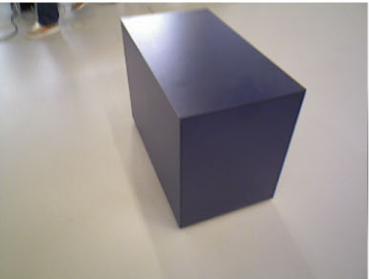}}
      \subfloat[\label{5b}]{%
          \includegraphics[width=0.29\linewidth]{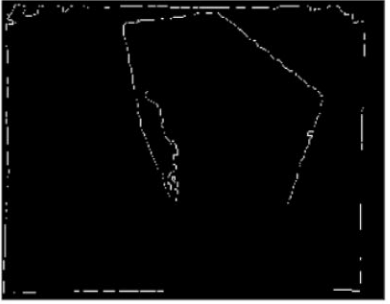}}
      \subfloat[\label{5c}]{%
          \includegraphics[width=0.295\linewidth]{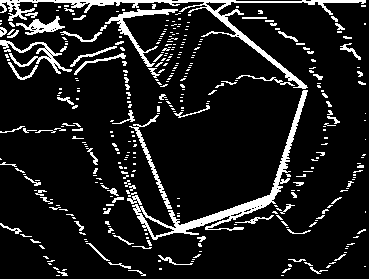}}
    \\
      \subfloat[\label{5d}]{%
          \includegraphics[width=0.3\linewidth]{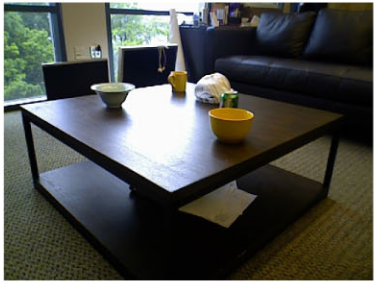}}
      \subfloat[\label{5e}]{%
          \includegraphics[width=0.3\linewidth]{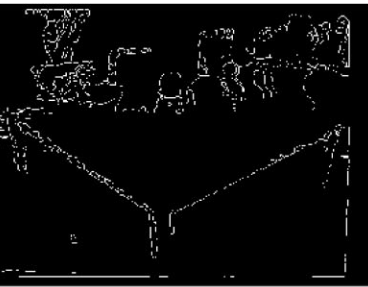}}
      \subfloat[\label{5f}]{%
          \includegraphics[width=0.3\linewidth]{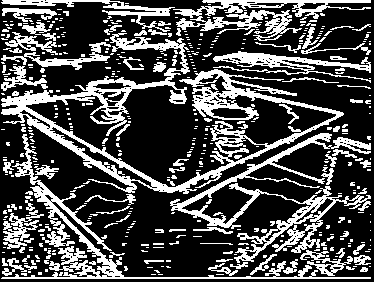}}
  
  \caption{(a), (d)  Input depth images in \cite{sung_depth_2020}, (b), (e) Edge detection output mask on a depth image in \cite{sung_depth_2020} (c), (f) edge detection output mask using pseudo-Boolean polynomials-based method on a depth image}
  \label{fig:pbp_thai}
  \end{figure}

\cite{sung_depth_2020} proposed incorporation of morphological properties to
the Canny edge detection method in order to detect edges in a
depth image.
In Fig.~\ref{fig:pbp_thai} we show that with our approach such
incorporation is not necessary since more edges are detected
and depths clearly preserved in the outputs

\section{Conclusion}

\noindent In this article in order to detect edge and blob regions in images we apply pseudo-Boolean polynomials to image patches. 
We show that our approach is fast, easy to explain and competitively accurate. We extend our approach towards semantic segmentation and optical character recognition based on its modification \cite{chikake_pseudo-boolean_2023}.
We invite our colleagues to find further applications since the power of this approach cannot be overestimated: in machine learning, deep neural
networks, and image recognition, among others \cite{umbaugh_digital_2011,prewitt_object_1970,li_pattern_2014,1211447,aroraMultilevelThresholdingImage2008}

\section*{Acknowledgments}

\noindent Tendai Mapungwana Chikake and Boris Goldengorin’s research was supported by Russian Science Foundation project No. 21-71-30005.

Boris Goldengorin acknowledges Scientific and Educational Mathematical Center ``Sofia Kovalevskaya Northwestern Center for Mathematical Research'' for financial support (agreement No 075-02-2023-937, 16.02.2023)

\vfill

\bibliographystyle{IEEEtran}
\bibliography{./bib}

\end{document}

\end{document}
\typeout{get arXiv to do 4 passes: Label(s) may have changed. Rerun}